\documentclass[11pt]{article}

\usepackage[preprint]{acl}

\usepackage{times}
\usepackage{latexsym}

\usepackage[T1]{fontenc}

\usepackage[utf8]{inputenc}

\usepackage{microtype}

\usepackage{inconsolata}

\usepackage{graphicx}

\usepackage[utf8]{inputenc} 
\usepackage[T1]{fontenc}    
\usepackage{hyperref}       
\usepackage{url}            
\usepackage{booktabs}       
\usepackage{amsfonts}       
\usepackage{nicefrac}       
\usepackage{microtype}      
\usepackage{xcolor}         
\usepackage{amsmath}
\usepackage{amssymb}
\usepackage{makecell}
\usepackage{graphicx}
\usepackage{subcaption}
\usepackage{multirow}
\usepackage{balance}
\usepackage{booktabs}
\usepackage[table]{xcolor}
\usepackage{makecell}
\definecolor{green1}{HTML}{d0f0c0}
\definecolor{green2}{HTML}{1ab200}
\definecolor{green3}{HTML}{6EC6C5}
\definecolor{green4}{HTML}{90EE90}
\usepackage[most]{tcolorbox}
\usepackage{xcolor}
\newtcolorbox{augprompt}{
  enhanced,
  colback=gray!5,
  colframe=green!70!black,
  colbacktitle=green!70!black,
  coltitle=white,
  fonttitle=\bfseries,
  title=Augmented Prompt,
  attach boxed title to top left={yshift=-2mm,xshift=3mm},
  boxed title style={
    boxrule=0pt,
  },
  arc=2mm,
  drop shadow,
}
\newtcolorbox{step2prompt}{
  enhanced,
  colback=gray!5,
  colframe=green!70!black,
  colbacktitle=green!70!black,
  coltitle=white,
  fonttitle=\bfseries,
  title=Extract Raw KPs,
  attach boxed title to top left={yshift=-2mm,xshift=3mm},
  boxed title style={
    boxrule=0pt,
  },
  arc=2mm,
  drop shadow,
}
\newcommand\blfootnote[1]{%
  \begingroup
  \renewcommand\thefootnote{}\footnote{#1}%
  \addtocounter{footnote}{-1}%
  \endgroup
}

\newtcolorbox{step3prompt}{
  enhanced,
  colback=gray!5,
  colframe=green!70!black,
  colbacktitle=green!70!black,
  coltitle=white,
  fonttitle=\bfseries,
  title=Leakage Verification,
  attach boxed title to top left={yshift=-2mm,xshift=3mm},
  boxed title style={
    boxrule=0pt,
  },
  arc=2mm,
  drop shadow,
}

\title{KnowRL: Boosting LLM Reasoning via Reinforcement Learning with Minimal-Sufficient Knowledge Guidance}

\author{
  Linhao Yu$^{1^*}$,
  Tianmeng Yang$^{2^*}$,
  Siyu Ding$^{2^*}$,
  Renren Jin$^1$,
  Naibin Gu$^{3}$,
  Xiangzhao Hao$^{3}$,\\
  \textbf{Shuaiyi Nie}$^{2}$,
  \textbf{Deyi Xiong$^{1^\ddagger}$,}
  \textbf{Weichong Yin$^{2}$,}
  \textbf{Yu Sun$^{2}$,}
  \textbf{Hua Wu$^{2}$}\\
  $^1$TJUNLP Lab, School of Computer Science and Technology, Tianjin University, China\\
  $^2$Baidu Inc.
  $^3$Institute of Information Engineering, Chinese Academy of Sciences. \\
  \texttt{\{linhaoyu, dyxiong\}@tju.edu.cn}, \texttt{\{yangtianmeng, dingsiyu\}@baidu.com}
}

\begin{document}
\maketitle
\blfootnote{
  $^*$Equal Contribution. \\
  \phantom{aliig}$^\ddagger$Corresponding author. \\
}
\begin{abstract}
  RLVR improves reasoning in large language models, but its effectiveness is often limited by severe reward sparsity on hard problems. Recent hint-based RL methods mitigate sparsity by injecting partial solutions or abstract templates, yet they typically scale guidance by adding more tokens, which introduce redundancy, inconsistency, and extra training overhead. We propose \textbf{KnowRL} (Knowledge-Guided Reinforcement Learning), an RL training framework that treats hint design as a minimal-sufficient guidance problem. During RL training, KnowRL decomposes guidance into atomic knowledge points (KPs) and uses Constrained Subset Search (CSS) to construct compact, interaction-aware subsets for training. We further identify a pruning interaction paradox---removing one KP may help while removing multiple such KPs can hurt---and explicitly optimize for robust subset curation under this dependency structure. We train KnowRL-Nemotron-1.5B from OpenMath-Nemotron-1.5B. Across eight reasoning benchmarks at the 1.5B scale, KnowRL-Nemotron-1.5B consistently outperforms strong RL and hinting baselines. Without KP hints at inference, KnowRL-Nemotron-1.5B reaches 70.08 average accuracy, already surpassing Nemotron-1.5B by +9.63 points; with selected KPs, performance improves to 74.16, establishing a new state of the art at this scale. The model, curated training data, and code are publicly available at \url{https://github.com/Hasuer/KnowRL}.
\end{abstract}

\section{Introduction}
RLVR has emerged as a paradigm for improving LLM reasoning by optimizing verifiable correctness \citep{DBLP:journals/corr/abs-2602-02276,DBLP:journals/nature/GuoYZSWZXZMBZY025,wang2026ernie50technicalreport, DBLP:journals/corr/abs-2505-09388, nie2026attnpoattentionguidedprocesssupervision}. By aligning outputs with rule-based verifiers, RLVR provides scalable supervision without relying on human preference annotations. However, RLVR suffers from a key bottleneck: \emph{reward sparsity} on difficult samples. For complex reasoning tasks, LLMs often produce uniformly incorrect rollouts, yielding zero advantage under group-based optimization methods such as GRPO \citep{shao2024deepseekmathpushinglimitsmathematical}. Consequently, a large portion of training data fails to contribute gradients, reducing learning efficiency.

To address this issue, recent work introduces \emph{hint-based RL}, which injects auxiliary guidance into prompts to increase the probability of generating reward-yielding responses. Existing approaches can be categorized into three types:
(1) \textbf{fixed-ratio solution-prefix hints} that prepend a partial reference solution using a predefined, constant hint ratio across training (e.g., QuestA \citep{DBLP:journals/corr/abs-2507-13266}, POPE \citep{qu2026popelearningreasonhard});
(2) \textbf{adaptive solution-based hints} that dynamically determine the hint ratio based on instance difficulty or training state (e.g., StepHint \citep{DBLP:journals/corr/abs-2507-02841}, UFT \citep{DBLP:journals/corr/abs-2505-16984}); and
(3) \textbf{abstraction-based hints} that provide reasoning templates or conceptual abstractions generated by teacher models (e.g., TAPO \citep{wu2025templaterlstructuredtemplateguidedreinforcement}, Guide \citep{DBLP:journals/corr/abs-2506-13923}, Scaf-GRPO \citep{DBLP:journals/corr/abs-2510-19807}).

Despite their differences, these methods implicitly treat hint design as a \emph{quantity expansion problem}, assuming that stronger guidance requires longer prefixes or richer abstractions. As a result, they largely overlook the issue of \emph{guidance redundancy}.

\begin{figure*}[t]
  \centering
  \begin{subfigure}[t]{0.25\linewidth}
    \centering
    \includegraphics[width=\linewidth]{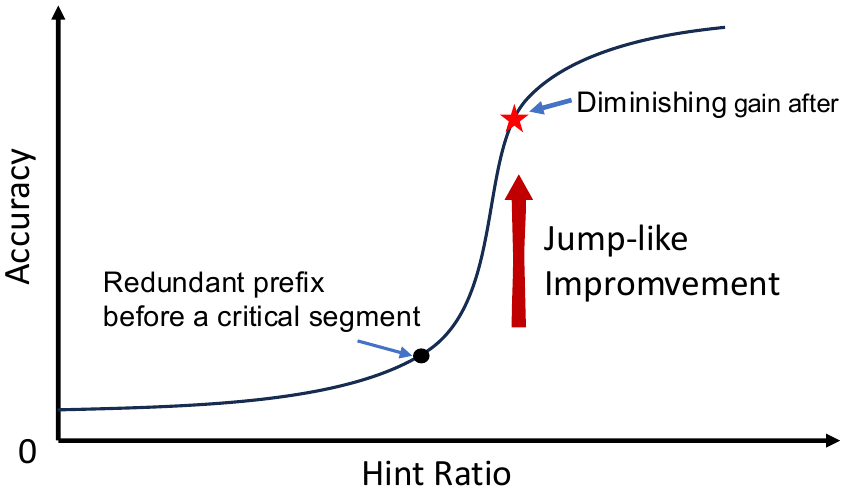}
    \caption{Critical-segment effect.}
    \label{fig:challenge1}
  \end{subfigure}
  \hfill
  \begin{subfigure}[t]{0.45\linewidth}
    \centering
    \includegraphics[width=\linewidth]{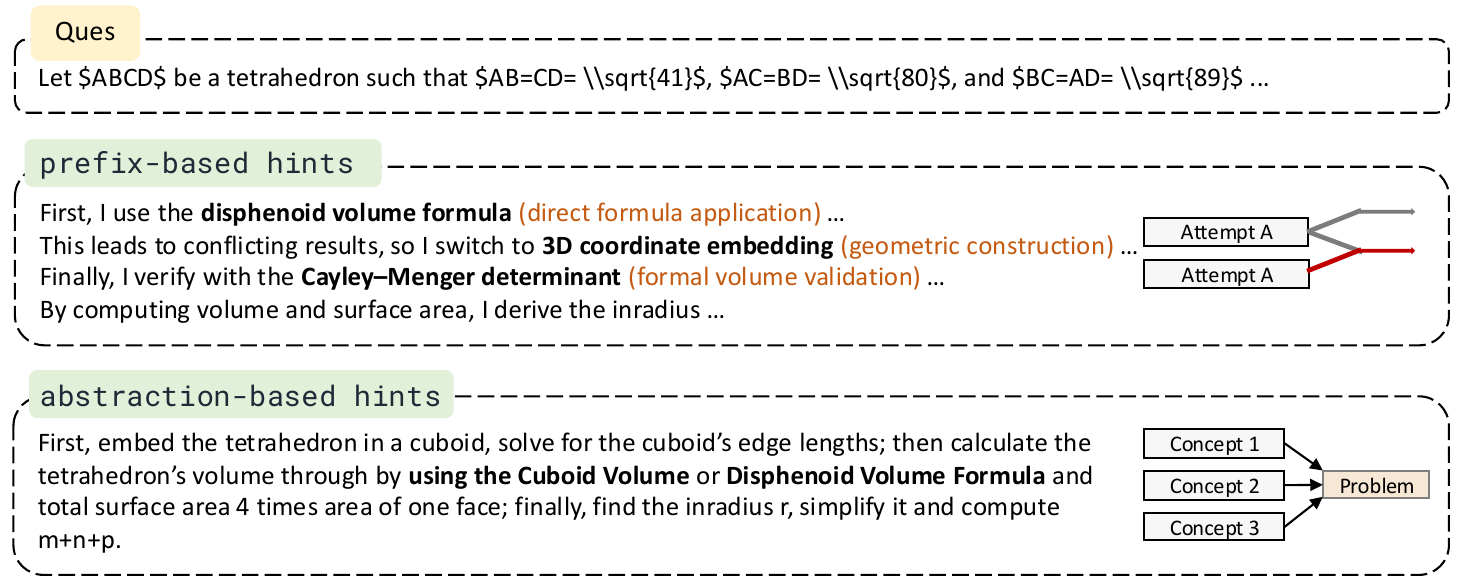}
    \caption{Cross-hint inconsistency.}
    \label{fig:challenge2}
  \end{subfigure}
  \begin{subfigure}[t]{0.25\linewidth}
    \centering
    \includegraphics[width=\linewidth]{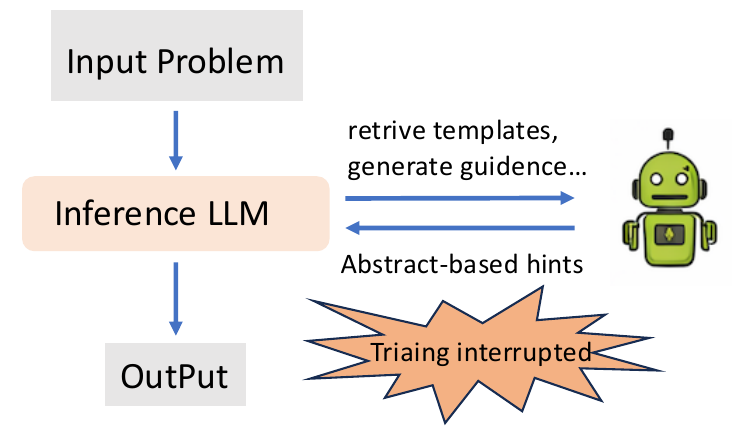}
    \caption{Guidance-efficiency trade-off.}
    \label{fig:challenge3}
  \end{subfigure}

  \caption{
    Three key challenges in hint-based RL.
    (a) Critical-segment effect: performance improves sharply once a short key hint segment appears, with diminishing returns beyond it.
    (b) Cross-hint inconsistency: longer prefixes or abstractions may introduce branching or ambiguity, expanding the reasoning search space.
    (c) Guidance-efficiency trade-off: abstraction-based hints often rely on teacher models or multi-stage curation, increasing computational overhead.
  }
  \label{fig:challenge}
\end{figure*}

Taken together, the three challenges above point to a shared pattern rather than three isolated drawbacks. In particular, they suggest that current hinting strategies often provide more guidance than is actually necessary, without sufficiently controlling its structure or relevance. We therefore argue that these limitations stem from a common issue: \textbf{hint redundancy}. Existing strategies often inject excessive or loosely structured guidance, while only a small subset of information is required to trigger successful reasoning.

First, we observe the \emph{critical-segment effect}: performance does not increase proportionally with hint ratio. Instead, accuracy exhibits a sharp jump once a short key segment appears, followed by diminishing gains (Figure~\ref{fig:challenge1}). This indicates that only a small set of knowledge components is sufficient to shift the policy toward reward-yielding trajectories. Appendix \ref{appendix:visualization} further visualizes this jump across varying hint ratios on 50 randomly sampled training instances. Second, we identify \emph{cross-hint inconsistency} (Figure~\ref{fig:challenge2}): longer prefixes or abstract templates may introduce branching and conceptual ambiguity, complicating policy updates. Third, we observe a trade-off between guidance independence and training efficiency (Figure~\ref{fig:challenge3}). Abstraction-based hints often rely on teacher-generated guidance, interrupting online RL and increasing computational cost.

Together, these findings suggest that the core challenge is not providing \emph{more} guidance, but selecting \emph{minimal, coherent knowledge units} that are sufficient to overcome reward sparsity. This naturally raises a fundamental question: \emph{can models be effectively trained using \emph{minimal yet sufficient hints} that unlock rewards without introducing redundant guidance?}

From an optimization perspective, the role of hints is not to replace reasoning but to shift the policy distribution toward reward-yielding trajectories. Therefore, the goal of hint design should be to provide the \emph{minimal information necessary to break reward sparsity}, rather than maximizing the amount of guidance injected into the prompt, as in most previous methods.

To this end, we propose \textbf{Knowledge-Guided Reinforcement Learning (KnowRL)}, a framework that formulates hint design as a \emph{minimal sufficient guidance problem}. Instead of injecting long solution prefixes or full reasoning templates, KnowRL decomposes guidance into atomic knowledge points (KPs) and identifies the minimal subset required to unlock reward learning. Importantly, we model a \emph{pruning interaction paradox}: removing a single KP may improve accuracy, while removing multiple such ``bad'' KPs together can reduce accuracy due to inter-KP dependencies. This paradox further guides our KP selection strategy design.

We construct KP hints for the training set through a structured pipeline and explore multiple KP selection strategies. Our final method adopts Constrained Subset Search (CSS), which prunes first and then performs global search over the remaining candidates, achieving the best performance with the fewest KPs. In practice, simple problems receive no hints, while minimal KP subsets are injected only for harder samples during training. The resulting RL-trained model achieves new state-of-the-art results across eight benchmarks with or without KP hints at inference, indicating that effective guidance depends on critical knowledge structure rather than long prefixes or heavy abstraction templates.

In summary, our contributions are threefold:

\begin{enumerate}
  \item \textbf{Minimal-sufficiency perspective on hint-based RL.} We introduce a minimal-sufficiency perspective on hint-based RL and empirically demonstrate a non-linear, jump-like performance pattern (critical-segment effect), revealing that effective guidance depends on selective key knowledge rather than cumulative hint length.

  \item \textbf{Principled KP selection pipelines.} We design several KP selection pipelines that ensure minimal, non-redundant, and interaction-compatible KP subsets for each problem. We further conduct detailed comparative analyses and finally identify CSS as the optimal selection strategy.

  \item \textbf{Efficient integration with state-of-the-art results.} We integrate minimal KP subsets into RL training via difficulty-aware prompt injection, achieving new state-of-the-art results across benchmarks while significantly reducing hint length and computational overhead.
\end{enumerate}

\section{Related Work}
\paragraph{Solution-Prefix Hints}
These methods typically extract fixed proportions of solution prefixes to guide the model. QuestA \citep{DBLP:journals/corr/abs-2507-13266} and Hint \citep{DBLP:journals/corr/abs-2510-09388} augment hard prompts with a fixed p\% solution prefix, while POPE \citep{qu2026popelearningreasonhard} further refines this approach by optimizing prefix selection based on token-level importance scores, but retains the core characteristic of fixed-ratio truncation.

\paragraph{Adaptive Solution-Based Hints}
To overcome the rigidity of fixed-ratio prefixes, subsequent works introduce adaptivity into solution-based hints, refining how and when guidance is injected. GHPO \citep{DBLP:journals/corr/abs-2507-10628}, G²RPO-A \citep{DBLP:journals/corr/abs-2508-13023} and Hint-GRPO \citep{DBLP:journals/corr/abs-2503-23905} scale hint length with task difficulty or recent reward signals, while StepHint \citep{DBLP:journals/corr/abs-2507-02841} refines granularity by partitioning reasoning chains into semantic steps for multi-level control. ADHint \citep{DBLP:journals/corr/abs-2512-13095} further incorporates offline difficulty priors to pre-calibrate hint strength, and DeepVideo-R1 \citep{DBLP:journals/corr/abs-2506-07464} extends this to video reasoning by coupling hint scaling with noise augmentation for simple samples.

Alongside these adaptive refinements, a parallel line of work incorporates solution prefixes into hybrid SFT--RL pipelines. BREAD \citep{DBLP:journals/corr/abs-2506-17211} ensures at least one successful trajectory per update by increasing the proportion of expert prefixes upon failure; Prefix-RFT \citep{DBLP:journals/corr/abs-2507-01679} concatenates offline SFT prefixes with online RL continuations to produce hybrid rollouts; and UFT \citep{DBLP:journals/corr/abs-2505-16984} employs a cosine-annealing schedule to progressively reduce hint length during training.

\paragraph{Abstraction-Based Hints}
Abstraction-based hints shift guidance from solution prefixes to high-level concepts, principles, and structured reasoning patterns rather than partial solutions. Guide \citep{DBLP:journals/corr/abs-2506-13923} introduces natural-language hints generated by stronger teacher models (e.g., GPT-4o) to accelerate learning on hard problems; Scaf-GRPO \citep{DBLP:journals/corr/abs-2510-19807} proposes a two-stage scaffold injection, where abstractions are generated by DeepSeek. Complementing this line, TAPO \citep{wu2025templaterlstructuredtemplateguidedreinforcement} incorporates structured “thought patterns” as external templates that encode general reasoning strategies.

More recently, abstraction generation itself has been incorporated into the learning objective. Self-Hinting \citep{liao2026selfhintinglanguagemodelsenhance} enables the model to act as its own teacher: given a solution, it generates abstract hints to guide subsequent rollouts, reducing reliance on external teachers. RLAD \citep{DBLP:journals/corr/abs-2510-02263} further refines this idea by training models with auxiliary supervision to produce higher-quality abstractions during RL. Complementary to hint-based RL, another line of work improves reasoning through distillation. Chen et al. \citep{chen-etal-2025-improving-reasoning} propose a CoT distillation framework that transfers the teacher's stepwise attention on key information to the student model, together with mixture-of-layers alignment for dynamic teacher--student matching. 

These methods typically depend on strong teacher models or carefully designed templates, and overly vague abstractions may fail to provide actionable signals for difficult reasoning tasks.

\section{KnowRL}

\begin{figure}[t]
  \centering
  \begin{subfigure}[t]{0.48\linewidth}
    \centering
    \includegraphics[width=\linewidth]{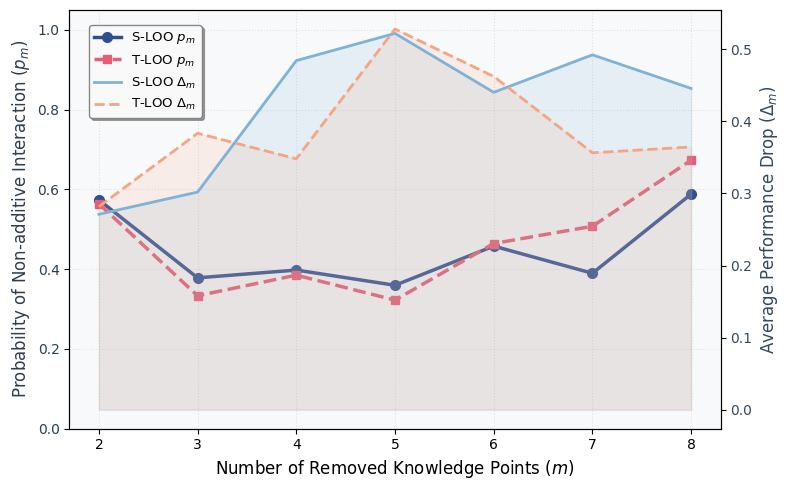}
    \caption{Pruning interaction paradox under LOO-style selection strategies.}
    \label{fig:subB}
  \end{subfigure}
  \hfill
  \begin{subfigure}[t]{0.48\linewidth}
    \centering
    \includegraphics[width=\linewidth]{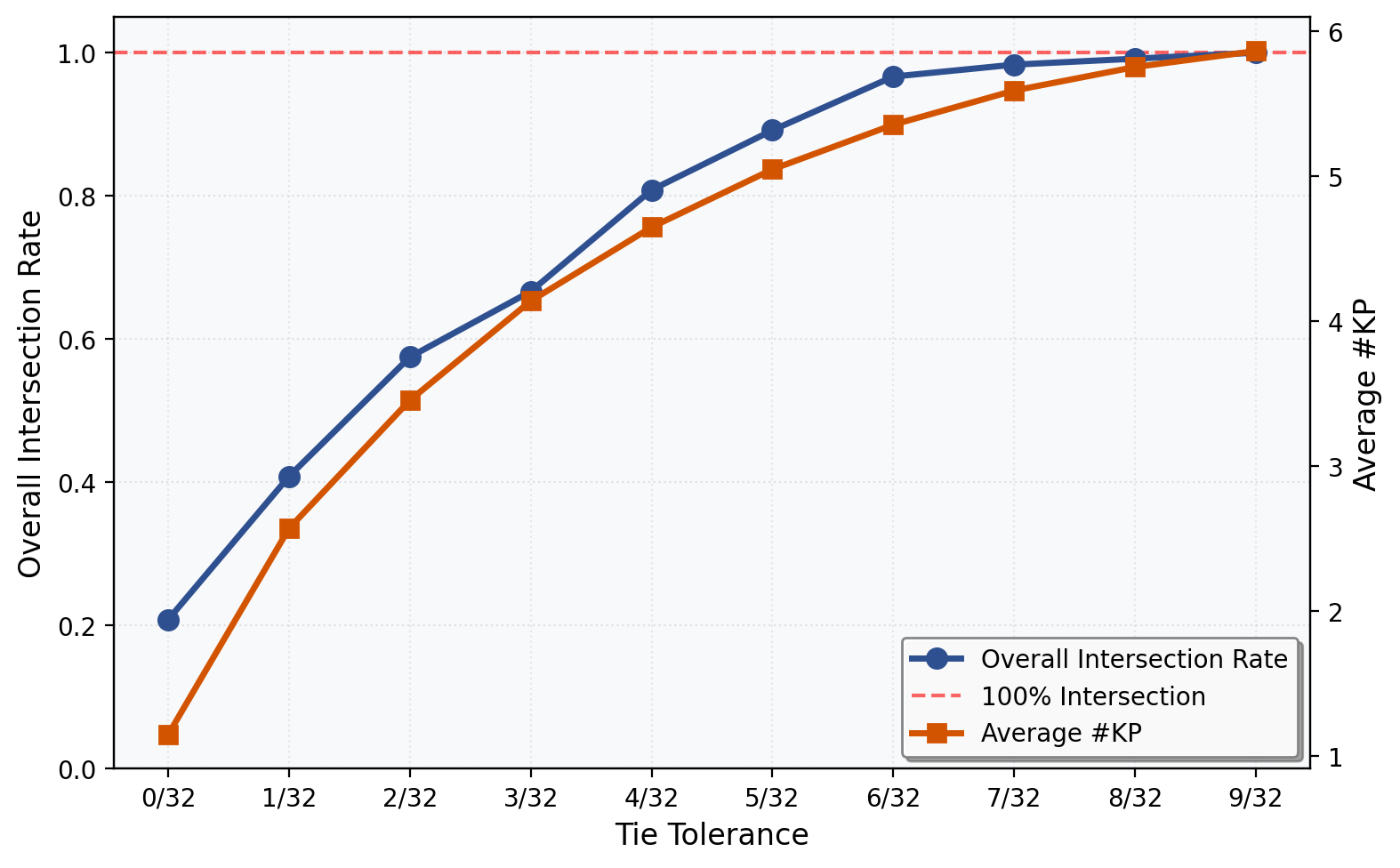}
    \caption{Tolerance-threshold sensitivity.}
    \label{fig:subD}
  \end{subfigure}

  \caption{Interaction-aware KP selection: inconsistency-induced degradation and the $\delta$--compactness trade-off.}
  \label{fig:three_ratio_figss}
\end{figure}

\begin{figure*}[t]
  \centering
  \begin{subfigure}[t]{0.48\linewidth}
    \centering
    \includegraphics[width=\linewidth]{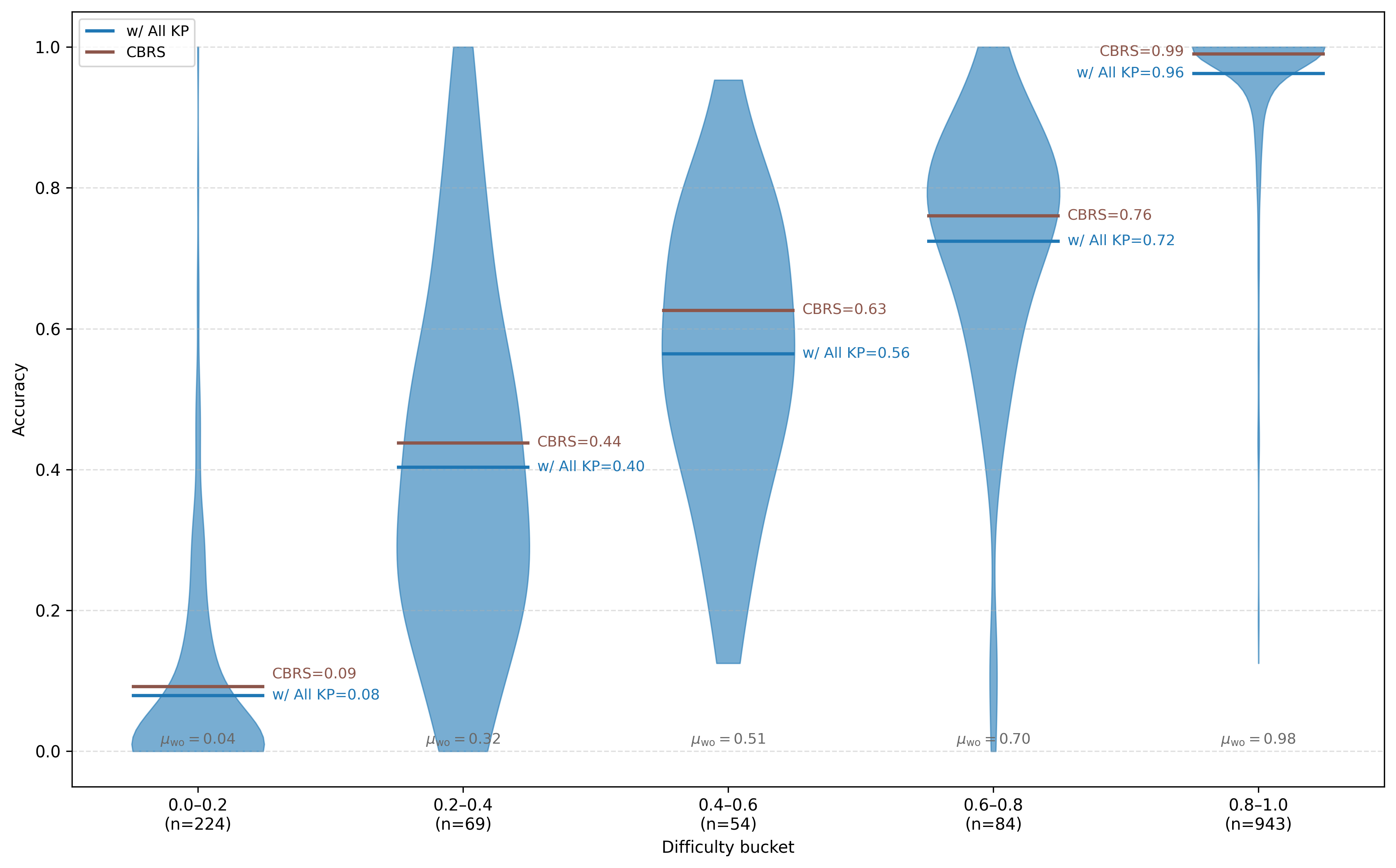}
    \caption{Test-set comparison across difficulty levels.}
    \label{fig:subC}
  \end{subfigure}
  \hfill
  \begin{subfigure}[t]{0.48\linewidth}
    \centering
    \includegraphics[width=\linewidth]{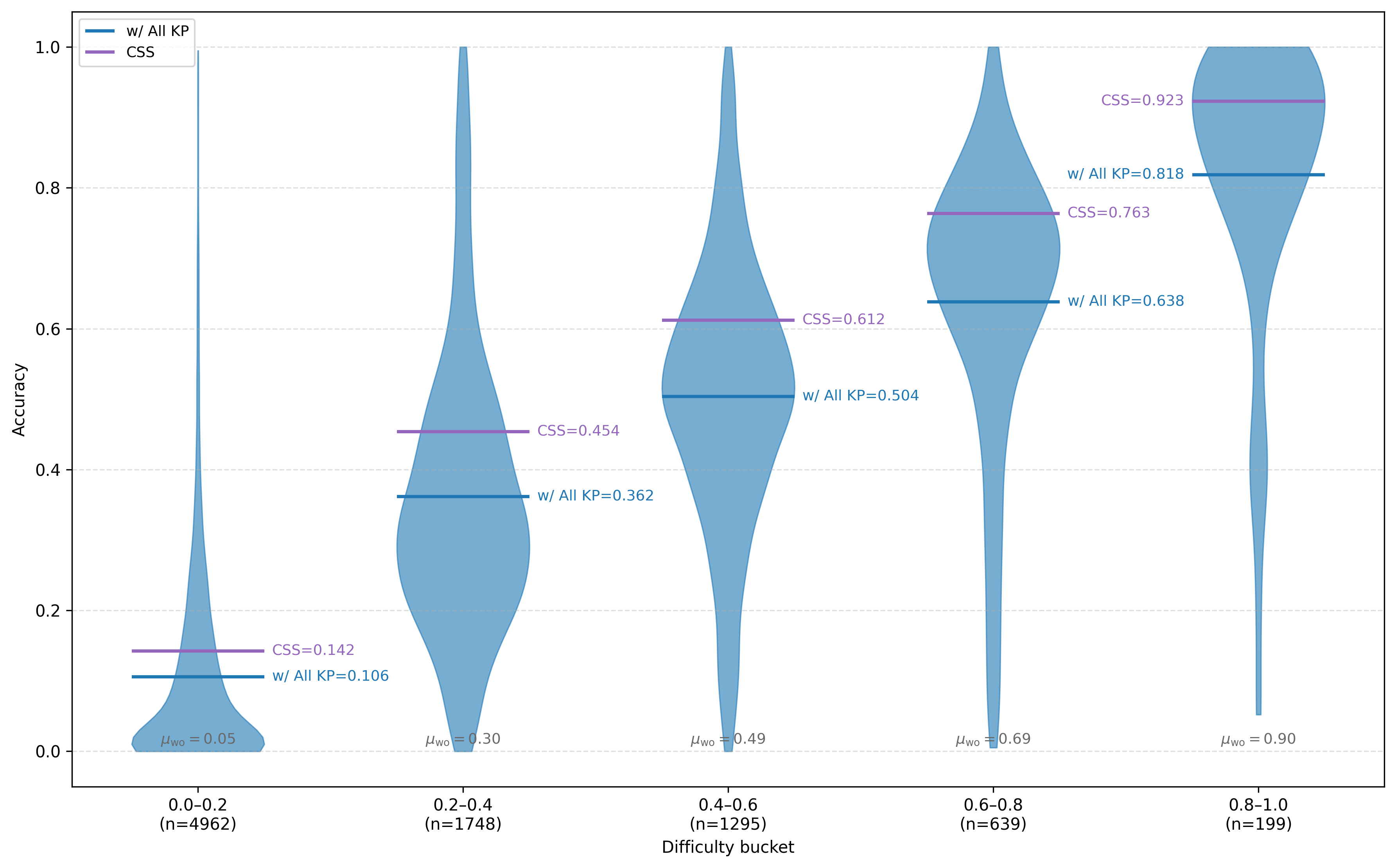}
    \caption{Training-set comparison across difficulty levels.}
    \label{fig:subE}
  \end{subfigure}

  \caption{Difficulty-bucket analysis on both test and training sets, where buckets are defined by no-KP accuracy. Full-KP injection shifts the violin distributions upward and improves mean performance in most buckets, but it also induces regressions on a subset of instances. In contrast, CSS-selected KPs deliver larger and more consistent gains across buckets. On the x-axis, $(n=\cdot)$ denotes the number of samples in each bucket, and the gray marker $\mu_{\mathrm{wo}}$ indicates the no-KP mean accuracy of that bucket.}
  \label{fig:three_ratio_figs}
\end{figure*}

In this section, we present KnowRL from a framework perspective. At a high level, KnowRL follows a simple end-to-end workflow: for each training problem, it first constructs candidate knowledge points (KPs), then removes leakage and redundancy to obtain a compact problem-specific subset, and finally uses the curated subset as hint data for RL training only when guidance is needed. In this sense, KnowRL is a complete training framework, but its central technical component is the construction of high-quality KP data.

Accordingly, this section focuses on the data-construction side of KnowRL, which is also the key component that determines the quality of the overall framework. We curate and analyze KP annotations over eight mathematical reasoning benchmarks: AIME24 \citep{aime24}, AIME25 \citep{aime25}, BRUMO25 \citep{balunovic_srimatharena_2025}, HMMT-Feb-25 \citep{balunovic_srimatharena_2025}, AMC23 \citep{li2024numinamath}, CMIMC25 \citep{balunovic_srimatharena_2025}, MATH-500 \citep{DBLP:conf/nips/HendrycksBKABTS21}, and Olympiad-Bench \citep{DBLP:conf/acl/HeLBHTSHHHZLQL024}, totaling 1,374 problems. All construction and selection procedures are performed via offline evaluation before RL training, ensuring reproducibility and computational efficiency.

\subsection{KP Curation}

The first stage of KnowRL is to construct candidate KP annotations for each problem through a three-stage pipeline. The prompts we use are shown in Appendix \ref{appendix:appendix_extract_kp_prompt}.

\textbf{Generating Correct Solutions.} For each problem, we sample responses from \texttt{DeepSeek-R1} until at least one correct solution is obtained. This guarantees that subsequent KP extraction is grounded in valid reasoning trajectories.

\textbf{Extracting Raw Knowledge Points.} Given a problem and a verified correct solution, we prompt \texttt{DeepSeek-R1} to extract only the indispensable mathematical principles required to solve the problem. This procedure yields an initial candidate KP set $\mathcal{K} = \{k_1, k_2, \dots, k_n\}.$

\textbf{Leakage Verification.} To prevent information leakage, we verify each KP using \texttt{DeepSeek-R1} as an automated reviewer. Failed cases are manually revised to ensure all retained KPs are generalizable and not instance-bound.

We evaluate OpenMath-Nemotron-1.5B with all KPs on 8 benchmarks. As shown in Table~\ref{tab:kp_selection}, the average performance improves from 60.46 to 61.03, but each problem uses 5.86 KPs on average. This result shows that raw KP construction alone is not sufficient: to make KnowRL effective as a complete framework, we also need a principled way to turn candidate KPs into compact training-ready hints. However, due to \emph{cross-hint inconsistency}, adding more KPs is not always better. We therefore study problem-wise KP subset selection as the second stage of the KnowRL data construction pipeline.

\subsection{Problem-wise KP Subset Selection}

For a problem with candidate KP set $\mathcal{K}$, we estimate offline accuracies under different configurations: $A_{\emptyset}$, $A_{\mathcal{K}}$, and $A_{-i} = A(\mathcal{K} \setminus \{k_i\})$.
Here, $A_{-i}$ corresponds to leave-one-out ablation of $k_i$, allowing us to quantify the marginal importance of each KP by measuring the performance drop when it is removed.
All accuracy estimates are computed using $8 \times 32$ samples to reduce variance.

A straightforward and practical strategy is ``Max-Score'', selecting the configuration achieving the highest accuracy among $\{\emptyset, \mathcal{K}, \mathcal{K}\setminus\{k_i\}\}.$ While effective (Table~\ref{tab:kp_selection}), this strategy restricts each problem to choosing from only three types of configurations: using no KPs, using the full set $\mathcal{K}$, or removing exactly one KP (i.e., an $n\!-\!1$ subset). This coarse search space can cause mismatches: problems that benefit from fewer KPs may be assigned the full set. Consequently, the resulting selections can be suboptimal.

\begin{table*}[t]
  \centering
  \scriptsize
  \setlength{\tabcolsep}{3pt}
  \resizebox{\textwidth}{!}{%
    \begin{tabular}{lllllllllll}
      \toprule
      \makecell[l]{Selection\\Strategy} & AIME24 & AIME25 & BRUMO25 & \makecell[l]{HMMT\\Feb 25} & AMC23 & CMIMC25 & MATH-500 & \makecell[l]{Olympiad\\Bench} & Avg. & \makecell[l]{Avg.\\\#KP}\\
      \midrule
      w/o KP & 58.75 & 48.44 & 61.67 & 30.10 & 90.55 & 30.08 & 92.40 & 71.70 & 60.46 & 0.00\\
      All KP & 60.90 & 49.01 & 61.11 & 32.46 & 89.67 & 32.32 & 92.22 & 70.55 & 61.03 & 5.86 \\
      Random & 60.52 & 49.27 & 61.04 & 33.23 & 91.02 & 31.09 & 91.65 & 71.88 & 61.21 & 2.53 \\
      Max-Score & 62.63 & 49.79 & 64.27 & 34.79 & 90.94 & 32.99 & 92.52 & 73.89 & 62.73 & 2.61 \\
      S-LOO & 62.71 & 49.22 & 63.88 & 33.54 & 91.71 & 33.52 & 92.90 & 73.70 & 62.65 & 1.72 \\
      T-LOO & 62.11 & 49.27 & 64.20 & 33.65 & 91.25 & 33.67 & 92.40 & 73.46 & 62.50 & 1.20 \\

      CBRS & 63.02 & 49.90 & 64.17 & 34.79 & 91.56 & 33.57 & 92.65 & 73.89 & 62.94 & 2.60 \\

      \rowcolor{gray!15}
      \textbf{CSS}
      & \textbf{64.44} {\tiny\textcolor{green2}{+5.69}}
      & \textbf{50.57} {\tiny\textcolor{green2}{+2.13}}
      & \textbf{65.03} {\tiny\textcolor{green2}{+3.36}}
      & \textbf{35.77} {\tiny\textcolor{green2}{+5.67}}
      & \textbf{91.71} {\tiny\textcolor{green2}{+1.16}}
      & \textbf{36.70} {\tiny\textcolor{green2}{+6.62}}
      & \textbf{92.90} {\tiny\textcolor{green2}{+0.50}}
      & \textbf{74.11} {\tiny\textcolor{green2}{+2.41}}
      & \textbf{63.90} {\tiny\textcolor{green2}{+3.44}}
      & 2.57\\
      \bottomrule
  \end{tabular}}
  \caption{Offline KP selection strategies on Nemotron-1.5B. Avg.\#KP denotes the average number of selected key knowledge points per problem. Green numbers indicate improvements over w/o KP.}
  \label{tab:kp_selection}
\end{table*}

\subsubsection{S-LOO and T-LOO}

We unify KP selection as a parameterized decision operator whose goal is to choose the most beneficial KP configuration for each problem, reducing dependence on KPs while preserving performance. To this end, we introduce a tolerance parameter $\varepsilon \ge 0$, which controls how strictly we treat borderline cases when selecting the optimal configuration.

Given $\varepsilon$, the generalized selection strategy is formalized as a mapping $\Phi_{\varepsilon} : \mathcal{K} \longrightarrow \mathcal{K}^* \subseteq \mathcal{K}$, where $\mathcal{K}^*$ denotes the final selected subset and is defined by

\[
  \Phi_{\varepsilon}(\mathcal{K}) =
  \begin{cases}
    \emptyset, & \text{if } A_{\emptyset} \ge \max(A_{\mathcal{K}},\, A_{\max}-\varepsilon), \\[6pt]

    \mathcal{K}, & \text{if } A_{\mathcal{K}} > \max(A_{\emptyset},\, A_{\max}-\varepsilon), \\[6pt]

    \mathcal{K} \backslash S, & \text{otherwise},
  \end{cases}
\]
with $S = \{k_i \mid A_{-i} < \max(A_{\mathcal{K}}, A_{\emptyset}) - \varepsilon\}$ and $A_{\max} = \max_i A_{-i}$.

Within this framework, different strategies correspond to different choices of $\varepsilon$. When $\varepsilon=0$, we obtain Strict Leave-One-Out selection (S-LOO). Since accuracy estimates are based on finite sampling and thus subject to randomness, we further introduce a tolerance band $\varepsilon = 1/32$, yielding Tolerant Leave-One-Out selection (T-LOO). Compared to S-LOO, T-LOO allows up to one-sample-scale performance rollback in near-tie cases, making selection more stable on borderline problems.

As shown in Table~\ref{tab:kp_selection}, S-LOO and T-LOO select substantially fewer KPs than Max-Score, but they also yield lower accuracy. A major reason is that LOO-based pruning overgeneralizes from single-KP ablations: even when removing $k_i$ alone improves accuracy, removing all such ``non-essential'' KPs together does not necessarily improve performance. In practice, this heuristic fails because of \emph{cross-hint inconsistency} and the pruning interaction paradox: KPs can be mutually dependent or implicitly disambiguate one another, so joint removal can introduce conflicts and cause larger-than-expected performance drops.

To quantify this effect, we characterize cases where removing each of $m$ KPs individually improves performance, but removing them jointly degrades performance. We define the positive-contribution set:
$\mathcal{K}^+ = \{k_i \mid A_{-i} \ge \max(A_{\mathcal{K}}, A_{\emptyset})\}.$
For subsets $S \subseteq \mathcal{K}^+$ with $|S|=m$, define: $A_{\text{joint}}(S) = A(\mathcal{K} \setminus S)$ and $\bar A_{\text{single}}(S) = \frac{1}{m}\sum_{k_i \in S} A_{-i}.$ Across problems, we compute: $p_m = \Pr(A_{\text{joint}}(S) < \bar A_{\text{single}}(S))$ and
$\Delta_m = \mathbb{E}[\bar A_{\text{single}}(S) - A_{\text{joint}}(S)| A_{\text{joint}}(S) < \bar A_{\text{single}}(S)].$ As summarized in Figure \ref{fig:subB}, cross-hint inconsistency occurs frequently (typically $p_m \in [40\%, 60\%]$), with substantial performance drops.

\subsubsection{Constrained Subset Search (CSS)}

To address this pruning interaction paradox, a theoretically optimal approach would evaluate all $2^n$ KP subsets, but this is computationally infeasible.

We instead construct a constrained search space. Define $\mathcal{H} = \{k_i \mid A_{-i} \ge \max(A_{\mathcal{K}}, A_{\emptyset})\}$ as the non-degrading KPs and $\mathcal{N} = \{k_i \in \mathcal{H} \mid A_{-i} \ge A_{\max}\}$ as near-optimal removals. KPs in $\mathcal{N}$ can be removed directly, since deleting them yields substantially improved performance. Moreover, since $|\mathcal{N}|$ is small on average (1.21), removing $\mathcal{N}$ alone rarely triggers the pruning interaction paradox.

Let $\mathcal{C} = \mathcal{H} \setminus \mathcal{N}$. We enumerate subsets only within $\mathcal{C}$, yielding search space size $2^{|\mathcal{C}|}$, which is tractable in practice. The final configuration is chosen via:
$S^* = \arg\max_S A(S)$,
over all constrained candidates plus $\emptyset$ and $\mathcal{K}$. As displayed in Table \ref{tab:kp_selection}, CSS achieves the best overall tradeoff: higher accuracy (63.90 eight-task average) with only 2.57 KPs per problem.

\subsubsection{Consensus-Based Robust Selection (CBRS)}

Instead of averaging $8 \times 32$ samples, CBRS treats each of the 8 runs independently.

For run $j$, define near-optimal configurations:
\[
  \mathcal{O}^{(j)} = \{c \mid A^{(j)}(c) \ge \max_{c'} A^{(j)}(c') - \delta\},
\]
with $\delta = 1/32$. We define robust consensus:
\[
  \mathcal{O}^* =
  \begin{cases}
    \bigcap_{j=1}^8 \mathcal{O}^{(j)}, & \text{if non-empty},\\
    \arg\max_c \sum_j \mathbf{1}(c \in \mathcal{O}^{(j)}), & \text{otherwise}.
  \end{cases}
\]

Further, when the above rules still yield multiple tied candidates, we select the one with the smaller score variance across the eight independent evaluation runs. Specifically, for any candidate configuration $c \in \mathcal O^\star$, let its performance variance over the eight evaluation runs be $\operatorname{Var}(c)
= \frac{1}{8}\sum_{j=1}^{8} \left(A^{(j)}(c) - \frac{1}{8}\sum_{j=1}^{8} A^{(j)}(c)\right)^2.$ We present the effect of selecting different $\delta$ values in Appendix \ref{appendix:appendix_delta}. As shown in Table \ref{tab:kp_selection}, CBRS also yields strong performance while maintaining compact KP sets.

\begin{table}[t]
  \centering
  \begin{tabular}{lll}
    \toprule
    \makecell[l]{KP Selection \\ Strategy} & Acc & Avg. \#KPs \\
    \midrule
    w/o KP & 22.40 & 0 \\
    w/ all KP & 26.93 {\small\textcolor{green2}{+4.53}} & 5.90 \\
    CBRS & 33.05 {\small\textcolor{green2}{+10.65}} & 3.68 {\small\textcolor{green2}{-37.7\%}} \\
    CSS & 33.51 {\small\textcolor{green2}{+11.11}} & 3.61 {\small\textcolor{green2}{-38.9\%}} \\
    \bottomrule
  \end{tabular}
  \caption{Offline evaluation on the QuestA dataset with different KP selection strategies and the average number of selected KPs.}
  \label{tab:rl_training_data}
\end{table}

\begin{table*}[t]
  \centering
  \setlength{\tabcolsep}{4pt}
  \resizebox{\textwidth}{!}{%
    \begin{tabular}{lllllllllll}
      \toprule
      Model & \makecell[l]{Hint\\Setting} & AIME24 & AIME25 & BRUMO25 & \makecell[l]{HMMT25} & AMC23 & CMIMC25 & MATH & \makecell[l]{OlyBench} & Avg. \\
      \midrule
      \multirow{3}{*}{Nemotron-1.5B \citep{DBLP:journals/corr/abs-2504-16891}}
      & w/o KP
      & 59.06 & 48.33 & 60.73 & 30.63 & 90.70 & 30.08 & 92.35 & 71.70 & 60.45 \\
      & CBRS
      & 63.02 & 49.00 & 64.17 & 34.79 & 91.56 & 33.57 & 92.65 & 73.89 & 62.94 \\
      & CSS
      & 64.06 & 50.10 & 65.03 & 35.77 & 90.47 & 36.70 & 92.90 & 74.09 & 63.64 \\
      \midrule
      \multirow{3}{*}{QuestA \citep{DBLP:journals/corr/abs-2507-13266}}
      & w/o KP
      & \textbf{71.56} & 62.08 & 67.5 & 40.94 & 93.44 & 41.48 & 92.95 & 72.28 & 67.78 \\
      & CBRS
      & 74.23 & 62.00 & 73.23 & 43.78 & 95.10 & 46.12 & 93.94 & 78.45 & 70.86 \\
      & CSS
      & 74.26 & 64.99 & 73.75 & 44.35 & 95.08 & 47.64 & 94.05 & 78.53 & 71.58 \\
      \midrule
      \multirow{3}{*}{JustRL \citep{DBLP:journals/corr/abs-2512-16649}}
      & w/o KP
      & 69.69 & 62.92 & 66.88 & 40.63 & \textbf{96.02} & 41.72 & 94.15 & 76.59 & 68.58 \\
      & CBRS
      & 69.76 & 62.36 & 70.49 & 41.81 & 95.7 & 44.45 & 94.85 & 78.41 & 69.73 \\
      & CSS
      & 70.42 & 61.43 & 70.67 & 41.54 & 95.54 & 45.19 & 94.59 & 78.68 & 69.76 \\
      \midrule
      \multirow{3}{*}{KnowRL-Nemotron-1.5B}
      & w/o KP
      & 69.79 {\tiny\textcolor{green2}{+10.73}} & \textbf{64.69} {\tiny\textcolor{green2}{+16.36}} & \textbf{69.48} {\tiny\textcolor{green2}{+8.75}} & \textbf{41.04} {\tiny\textcolor{green2}{+10.41}} & 95.55 {\tiny\textcolor{green2}{+4.85}} & \textbf{44.14} {\tiny\textcolor{green2}{+14.06}} & \textbf{95.70} {\tiny\textcolor{green2}{+3.35}} & \textbf{80.23} {\tiny\textcolor{green2}{+8.53}} & \textbf{70.08} {\tiny\textcolor{green2}{+9.63}} \\
      & CBRS
      & \textbf{75.52} {\tiny\textcolor{green2}{+12.50}} & \textbf{65.00} {\tiny\textcolor{green2}{+16.00}} & \textbf{78.33} {\tiny\textcolor{green2}{+14.16}} & \textbf{45.00} {\tiny\textcolor{green2}{+10.21}} & \textbf{95.78} {\tiny\textcolor{green2}{+4.22}} & \textbf{49.22} {\tiny\textcolor{green2}{+15.65}} & \textbf{96.45} {\tiny\textcolor{green2}{+3.80}} & \textbf{82.34} {\tiny\textcolor{green2}{+8.45}} & \textbf{73.46} {\tiny\textcolor{green2}{+10.52}} \\
      & CSS
      & \textbf{74.58} {\tiny\textcolor{green2}{+10.52}} & \textbf{65.21} {\tiny\textcolor{green2}{+15.11}} & \textbf{78.12} {\tiny\textcolor{green2}{+13.09}} & \textbf{48.75} {\tiny\textcolor{green2}{+12.98}} & \textbf{95.70} {\tiny\textcolor{green2}{+5.23}} & \textbf{52.19} {\tiny\textcolor{green2}{+15.49}} & \textbf{96.20} {\tiny\textcolor{green2}{+3.30}} & \textbf{82.44} {\tiny\textcolor{green2}{+8.35}} & \textbf{74.16} {\tiny\textcolor{green2}{+10.52}} \\

      \bottomrule
  \end{tabular}}
  \caption{Evaluation results of RL training with CSS-selected KP data under different test-time prompting strategies (with and without KPs). All scores are evaluated using the protocol described in Section~\ref{sec:evaluation}. For QuestA and JustRL, the w/ KP scores are taken directly from the JustRL paper.}
  \label{tab:hint_settings}
\end{table*}

\paragraph{Summary.}
While full-KP injection can improve performance, naive pruning strategies such as Max-Score or LOO often fail due to cross-hint inconsistency and the pruning interaction paradox. CBRS independently aggregates rollouts from multiple generation rounds, but it does not fully resolve this problem. In contrast, CSS further mitigates it by first pruning candidates and then conducting a global search over the pruned candidate space. Although CSS and CBRS select a similar number of knowledge points (around 2.5 on average), their Jaccard similarity is 0.70, indicating substantial but still incomplete overlap and thus clear strategy-specificity in the selected KP configurations. Figures~\ref{fig:subC} and \ref{fig:subE} compare full-KP input with CSS-selected KPs on the test and training sets, showing that the model achieves larger gains across different difficulty levels after selection. In contrast, injecting all KPs can even introduce negative effects on certain subsets, highlighting the importance of interaction-aware KP selection. To further isolate the effect of selection quality from the number of hints, we construct a random-KP baseline by sampling 2–3 knowledge points per problem (average $\approx$2.5), matching the cardinality of CSS. Offline evaluation shows that randomly selected KPs perform substantially worse than both CSS and CBRS, demonstrating that the effectiveness of hinting depends not merely on the number of knowledge points but critically on robust, interaction-aware selection. These findings directly motivate the KP selection pipeline used in our final RL training.

\section{Experiments}
In this section, we examine KnowRL from four aspects: training data construction, training setup, evaluation protocol, and final performance.
\subsection{Training Data}
We used the open-source QuestA dataset \citep{DBLP:journals/corr/abs-2507-13266}. We retained 8.8k training instances after deduplication.
For each instance, we sampled 32 generations with top\_p$=0.9$ and temperature $T=0.9$, and repeated this procedure over 8 independent runs. Following Section~3, we obtained KPs using the CSS strategy since it yielded more compact KP sets and the best offline performance. The post-processed KP statistics are reported in Table~\ref{tab:rl_training_data}, reducing the number of KPs by around 38\%.

\begin{figure*}[t]
  \centering
  \includegraphics[width=\linewidth]{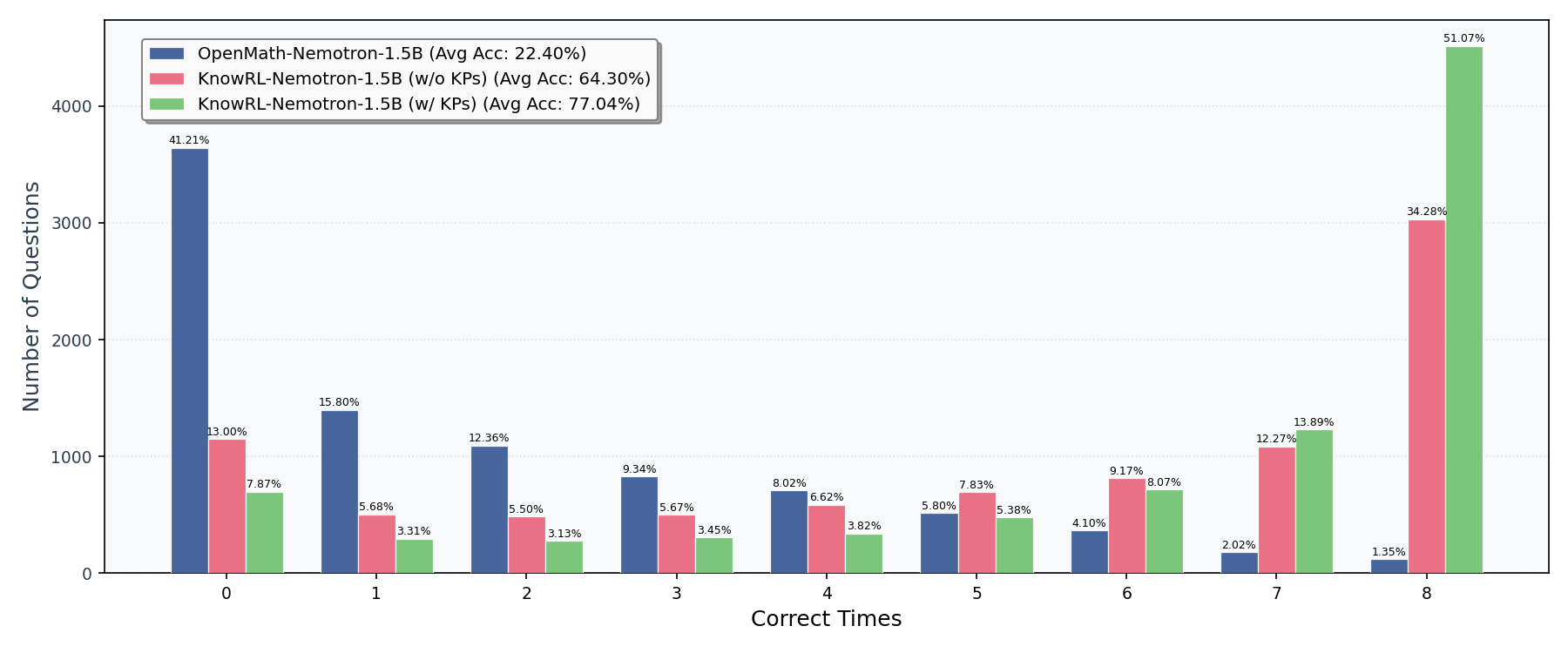}
  \caption{Distribution of per-query correct counts on the training set for OpenMath-Nemotron-1.5B, and KnowRL-Nemotron-1.5B model under two offline evaluation settings: without KP hints and with KP hints at inference.}
  \label{fig:cdf}
\end{figure*}

\subsection{Training Setup}
We set train\_batch\_size$=256$, performed four updates per step, and used a constant learning rate of $10^{-6}$ with clip\_ratio\_range $\in [0.8, 1.28]$. Each question was sampled eight times with top\_p$=1.0$ and $T=1.0$, and max\_response\_length was set to 24k. We used token-mean loss, did not use KL loss or an entropy bonus, and enabled dynamic sampling \citep{DBLP:journals/corr/abs-2503-14476}. We added pre-curated KPs to prompts under the \texttt{\#\# Hint} header; an example augmented prompt is provided in Appendix~\ref{appendix:appendix_augprompt}.

All experiments were conducted on a cluster of eight NVIDIA H100 nodes, each equipped with 8 GPUs. Training KnowRL-Nemotron-1.5B required approximately 13 days of wall-clock time. We used entropy annealing during training: with clip\_high $=0.28$, entropy increased early on (encouraging exploration), then began to decrease at step 2,590 as the model searched for optimal paths; to further accelerate convergence, following the findings of \citet{jin2026revisitingentropyreinforcementlearning}, we reduced clip\_high to $0.26$ after step 2,590. We additionally conducted a comparison experiment and reported the difference between using and not using annealing in Appendix \ref{appendix:appendix_entropy_annealing}.

\subsection{Evaluation Setup}
\label{sec:evaluation}
During training, we used a purely rule-based reward. For offline evaluation, we followed the JustRL-style protocol: first applied the rule-based evaluator based on \texttt{mathverify==0.8.0}, and when it failed, further verified with CompassVerifier-3B \citep{DBLP:conf/emnlp/LiuLLXGLGZWZC25}. We used a maximum length of 32k tokens, top\_p$=0.7$, and $T=0.9$, with 8 samples per problem on MATH-500 and Olympiad-Bench (reported as mean@8) and 32 samples per problem on the remaining benchmarks (reported as mean@32).

\subsection{Experiment Results}
On our carefully curated training set, we train OpenMath-Nemotron-1.5B \citep{DBLP:journals/corr/abs-2504-16891} for 2,960 steps and achieve a new state-of-the-art average accuracy of 70.08.
\begin{figure}[t]
  \centering
  \includegraphics[width=\linewidth]{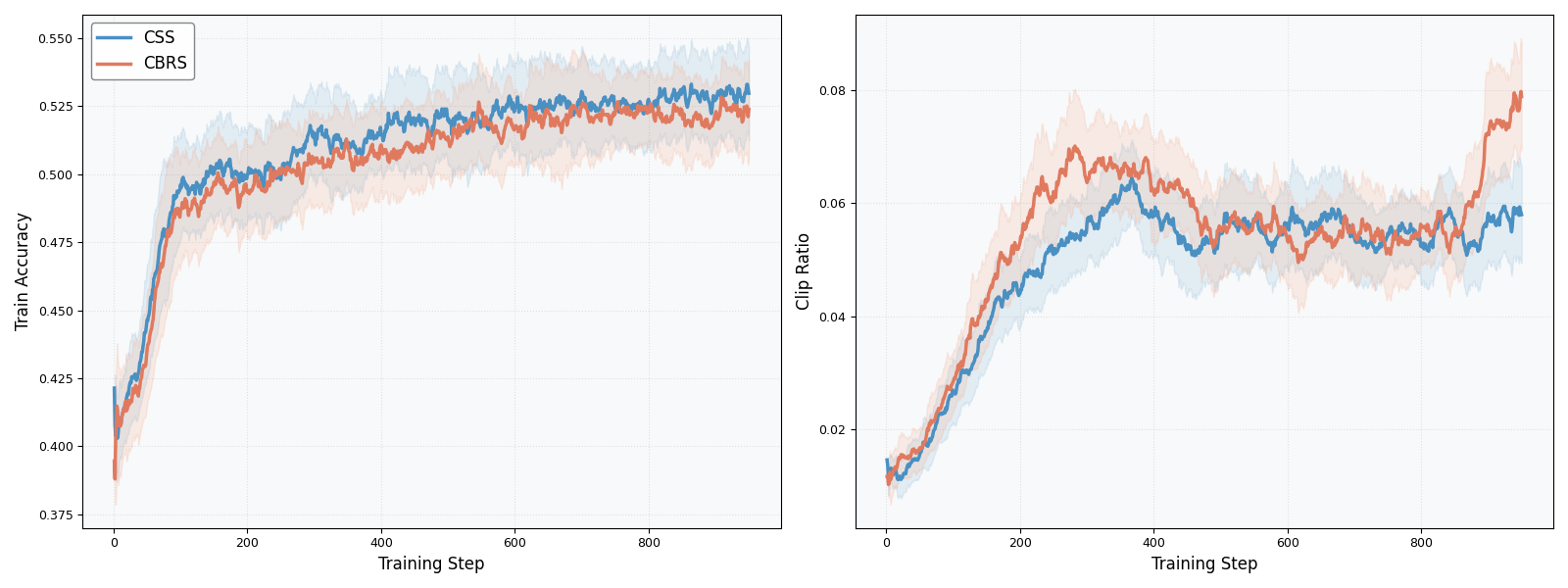}
  \caption{Comparison of KP selection strategies under the same training budget.}
  \label{fig:s2_select_strategy}
\end{figure}
\paragraph{Overall performance.}
Across all eight benchmarks, KnowRL-Nemotron-1.5B consistently achieves the strongest overall performance. Even without KP hints, KnowRL-Nemotron-1.5B reaches an average score of 70.08, already clearly surpassing Nemotron-1.5B by +9.63 points and outperforming JustRL by +1.50. When incorporating selected KPs, performance further improves to 73.46 with CBRS and 73.47 with CSS, establishing a new state of the art at the 1.5B scale. Notably, the substantial no-KP improvement (70.08) shows that KnowRL improves the underlying policy itself, rather than relying only on test-time hint injection.

\begin{table*}[t]
  \centering
  \setlength{\tabcolsep}{4pt}
  \resizebox{\textwidth}{!}{%
    \begin{tabular}{llllllllll}
      \toprule
      Model & AIME24 & AIME25 & BRUMO25 & \makecell[l]{HMMT25} & AMC23 & CMIMC25 & MATH & \makecell[l]{OlyBench} & Avg. \\
      \midrule
      CBRS {\tiny{step400}}  & 64.58 & 56.77 & 63.96 & 34.48 & 93.52 & 35.39 & 93.30 & 75.41 & 64.68 \\
      CSS {\tiny{step400}} & 65.94 & 57.08 & 64.38 & 35.31 & 92.73 & 36.25 & 92.85 & 75.46 & 65.00 \\
      CBRS {\tiny{step900}} & 65.42 & 58.85 & 64.79 & 37.08 & 94.14 & 35.70 & 94.00 & 75.78 & 65.72 \\
      CSS {\tiny{step900}} & 67.19 & 59.17 & 65.52 & 39.06 & 93.91 & 37.03 & 93.77 & 76.04 & 66.46 \\
      \bottomrule
  \end{tabular}}
  \caption{Comparison between CBRS- and CSS-selected training data under matched training budgets (steps 400 and 900) across eight reasoning benchmarks.}
  \label{tab:kp_select_acc_compare}
\end{table*}

The gains are particularly pronounced on more challenging competition-style reasoning benchmarks. Under CSS selection, KnowRL-Nemotron-1.5B achieves substantial improvements over Nemotron-1.5B without KP, including +15.11 on AIME25, +12.98 on HMMT25, and +15.49 on CMIMC25. These large margins suggest that interaction-aware KP selection effectively enhances long-horizon and compositional reasoning, rather than merely providing superficial guidance.

\paragraph{Selection strategy matters.}
Comparing the two selection strategies during offline evaluation, both CSS and CBRS consistently outperform vanilla training, but CSS shows stronger robustness on the hardest datasets, such as HMMT25 and CMIMC25, indicating that conflict-aware and interaction-sensitive selection leads to more reliable hint construction. Notably, KnowRL-Nemotron-1.5B also achieves leading performance on broader evaluation sets, reaching 96.20 on MATH-500, 82.44 on OlyBench, and 95.70 on AMC23, demonstrating that the improvements generalize across diverse reasoning distributions rather than being confined to a specific benchmark type.

Results validate that carefully selected KPs provide more effective training signals than both naive training and conventional hinting strategies, substantially improving reasoning performance while maintaining efficiency. Results indicate that KnowRL improves policy quality itself, rather than merely exploiting prompt-time scaffolding.

\paragraph{Improvements on Training Data.}
To further characterize KnowRL's effect, we analyze the per-query correct-count distribution (out of 8 samples) over the training set across three conditions, as shown in Figure~\ref{fig:cdf}.

The backbone suffers severely from reward sparsity: 41.21\% of queries receive zero correct answers and only 1.35\% are solved consistently, yielding a mean accuracy of 22.40\%. KnowRL training alone (w/o KPs at inference) collapses the zero-correct fraction to 13.00\% and raises the all-correct bucket to 34.28\% (+32.93pp), lifting average accuracy to 64.30\%. This confirms that KP-guided training genuinely internalizes structured reasoning rather than producing hint-conditioned shortcuts. Adding KP hints at inference further concentrates mass at the rightmost bucket (51.07\%), with mid-range counts (1–6) each shrinking by 2–3 percentage points, consistent with the critical-segment effect: once minimal sufficient knowledge is made explicit, the model resolves partial successes into consistent correctness. Average accuracy reaches 77.04\% under this condition.

\section{Comparison of KP Selection Strategies}
To further validate the training effectiveness of CSS-selected data, we compare CSS and CBRS under the same training budget, as shown in Figure~\ref{fig:s2_select_strategy}.
Both strategies select a comparable number of KPs per problem, enabling a fair comparison of selection quality rather than guidance quantity.

\textbf{Training Accuracy.}
CSS consistently achieves higher training accuracy throughout most of the optimization trajectory.
Although both methods rapidly improve during the first 200 steps, CSS maintains a persistent advantage and converges to a slightly higher final accuracy.

\textbf{Clip Ratio.}
CBRS exhibits a noticeably higher clip ratio during mid-to-late training and shows a sharp increase near the end of optimization.
In contrast, CSS maintains a smoother and more controlled clip ratio trajectory.
This indicates that CBRS induces more aggressive policy updates, while CSS leads to more stable policy refinement.

\textbf{Performance.}
As shown in Table~\ref{tab:kp_select_acc_compare}, CSS consistently generalizes better than CBRS under different training budgets: at the earlier checkpoint, CSS reaches 65.00 vs. 64.68 for CBRS, and at step 900, CSS further leads with 66.46 vs. 65.72. This trend supports the mechanism discussed in Section~3.3 and Section~3.4: CSS first prunes low-value candidates and then performs broader constrained enumeration, enabling a more thorough search for high-quality, global KP configurations; in contrast, CBRS relies on consensus among a relatively limited candidate pool, which is robust but can miss strong yet lower-frequency combinations.

\section{Conclusion}
We have presented KnowRL, a minimal-sufficient guidance framework for RLVR that decomposes hints into atomic knowledge points and selects robust subsets. Besides, we identify a jump-like critical-segment phenomenon and design a highly effective KP selection strategy, CSS, which explicitly handles inter-KP interactions and consistently outperforms alternatives while keeping hint sets compact. Across eight math reasoning benchmarks under matched budgets, KnowRL improves optimization stability and generalization, achieving a new 1.5B-scale state of the art. These results position compact, structured guidance as a practical scaling principle for sparse-reward RL, and motivate extending KP curation and robust selection to broader reasoning domains.

\bibliography{custom}

\clearpage
\appendix
\section{Visualization of the Critical-segment Effect.}
\label{appendix:visualization}

\begin{figure*}[t]
  \centering
  \includegraphics[width=0.95\linewidth]{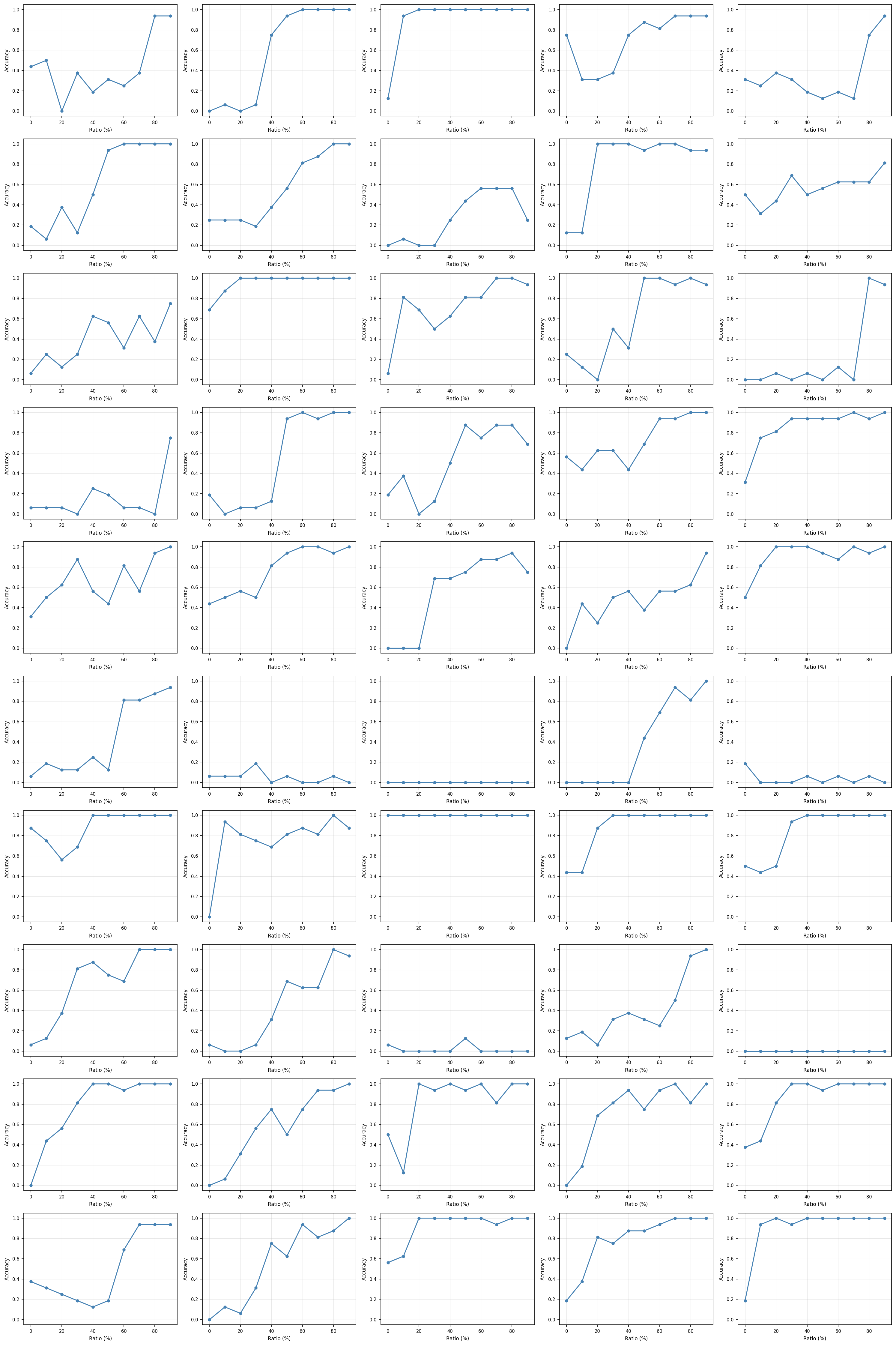}
  \caption{Visualization of the critical-segment effect across prefix ratios on 50 training instances.}
  \label{fig:visualization}
\end{figure*}

To further visualize the critical-segment effect, we conduct a controlled prefix-ratio study on the QuestA dataset and randomly choose 100 training instances for visualization. For each instance, we take the reference solution and append only the first $r\%$ prefix (with $r$ varying from 0 to 90) to the prompt followed by \texttt{\#\# Hint}, while keeping all other decoding and evaluation settings fixed. Figure~\ref{fig:visualization} shows that, for most instances, accuracy does not increase linearly as the injected prefix becomes longer. Instead, performance typically remains flat in low-ratio regions and then exhibits a distinct jump once a key segment is included, followed by diminishing gains. This pattern supports our view: effective guidance depends on whether critical knowledge is covered, rather than on monotonically increasing hint length.

\begin{figure*}[t]
  \centering
  \includegraphics[width=\linewidth]{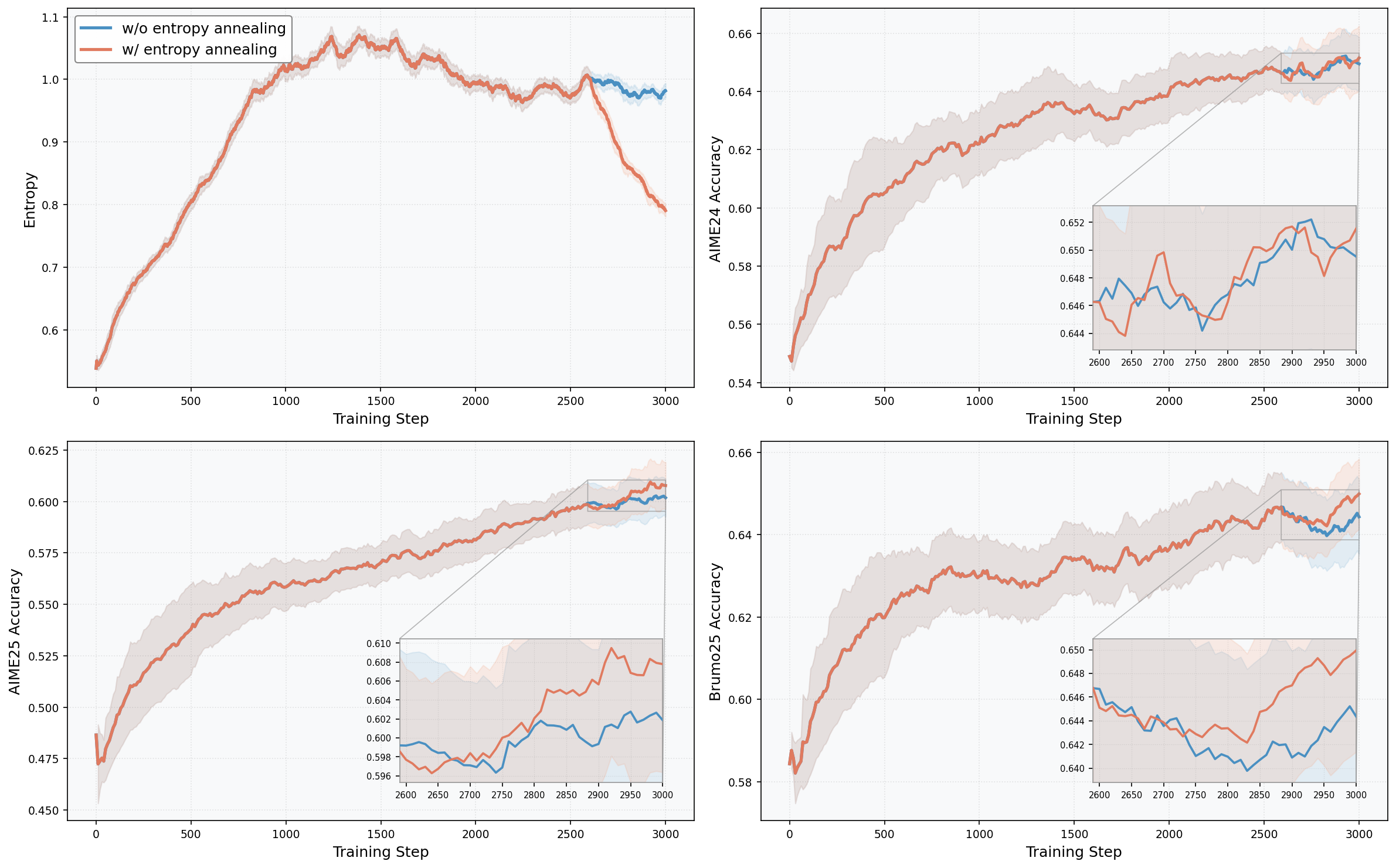}
  \caption{Comparison with and without entropy annealing. We report the entropy trajectory and the corresponding performance on different validation benchmarks under the 24k setting during training. Entropy annealing yields faster entropy reduction and consistently better validation performance.}
  \label{fig:entropy_anneal}
\end{figure*}

\section{Entropy Annealing Analysis}
\label{appendix:appendix_entropy_annealing}

To accelerate convergence under a limited training budget, we apply entropy annealing by adjusting the clip upper bound during training. Specifically, after 2{,}590 steps, we reduce \texttt{clip\_high} from 0.28 to 0.26. This tighter clipping regime induces a faster entropy drop, encouraging the policy to shift earlier from exploration to exploitation, which helps the model reach stronger performance within fewer optimization steps.

To isolate the contribution of this strategy, we compare against a control setting that keeps \texttt{clip\_high}=0.28 throughout training and runs to the same 2{,}960-step budget. Table~\ref{tab:entropy_annealing} further reports the detailed results on eight evaluation benchmarks, where the entropy-annealed setting achieves better overall scores than both the non-annealed variant and JustRL.

\section{Prompts}
\subsection{Prompts for KP Curation Pipeline}
\label{appendix:appendix_extract_kp_prompt}

In this section, we provide detailed prompt examples for the latter two stages introduced in Section~3.1, namely ``extracting raw knowledge points'' and ``leakage verification'', shown in Figure~\ref{fig:prompt_extract_raw_kp} and Figure~\ref{fig:prompt_verification_leakage}, respectively.

\begin{figure}[t]
  \centering
  \begin{step2prompt}
    You will be given: \\
    1. A mathematics problem.\\
    2. A correct solution to the problem.\\

    Your task is to extract the essential mathematical knowledge required to solve the problem from the correct solution.

    Requirements:\\
    - List only the core knowledge points that are indispensable for solving the problem.\\
    - Each knowledge point should be concise, general, and mathematically fundamental (not problem-specific tricks), yet crucial to the solution.\\
    - Do NOT explain the full solution or reproduce the reasoning steps.\\
    - Do NOT restate or paraphrase the problem.\\
    - For each knowledge point, explicitly describe important details, conditions, or caveats that must be considered when applying it.\\

    Format:\\
    - Use a numbered list.\\
    - Each item should consist of:\\
    (a) the knowledge point, and\\
    (b) the key considerations when applying it.\\

    [Problem]\\
    \texttt{\{question\}}

    [Correct Solution]\\
    \texttt{\{solution\}}

    [Key Knowledge Points]\
  \end{step2prompt}
  \caption{Prompt used for extracting raw knowledge points.}
  \label{fig:prompt_extract_raw_kp}
\end{figure}

\begin{figure}[t]
  \centering
  \begin{step3prompt}
    You are an expert reviewer for mathematical reasoning datasets.

    You will be given:\\
    1. A mathematics problem.\\
    2. A candidate knowledge description.\\

    Your task is to determine whether the knowledge description is STRONGLY COUPLED to the problem.

    Strong Coupling means the knowledge description:\\
    - Contains specific numerical values, constants, or quantities appearing in the problem or its solution (beyond common constants like $\pi$).\\
    - Mentions specific object names, variable names, or configurations unique to the problem.\\
    - Includes intermediate numerical results or the final answer.\\
    - Encodes the problem’s structure implicitly (e.g., ``in this case'', ``here'', ``for this triangle'', ``for this sequence'').

    If ANY of the above holds, the knowledge description IS strongly coupled.

    Respond STRICTLY in JSON format:\\
    \{\{\\
        ``strongly\_coupled'': true \/ false,\\
        ``reason'': ``<brief explanation>''\\
    \}\}\\

    [Problem]\\
    \texttt{\{question\}}

    [Knowledge Description]\\
    \texttt{\{knowledge\}}.
  \end{step3prompt}
  \caption{Prompt used for leakage verification with an augmented hint.}
  \label{fig:prompt_verification_leakage}
\end{figure}

\begin{figure}[t]
  \centering
  \begin{augprompt}
    Jackson's paintbrush makes a narrow strip with a width of $6.5$ millimeters. Jackson has enough paint to make a strip $25$ meters long. How many square centimeters of paper could Jackson cover with paint?

    \begin{tcolorbox}[
        colback=green1!35,
        colframe=green2!70!black,
        boxrule=0pt,
        arc=1mm,
        left=1mm,
        right=1mm,
        top=0.5mm,
        bottom=0.5mm
      ]
      \textbf{\#\# Hint}\\
      1. \textbf{Knowledge Point}: Unit conversion between metric length units follows powers of 10 (millimeters, centimeters, meters, etc.).\\
      \textbf{Key Considerations}: Moving to a larger unit requires division; moving to a smaller unit requires multiplication. The conversion factors are: 10 mm \textnormal{=} 1 cm, 100 cm \textnormal{=} 1 m.\\
      2. \textbf{Knowledge Point}: When calculating area, all linear measurements must be converted to the same unit before computation.\\
      \textbf{Key Considerations}: The choice of unit should match the desired unit of the final answer; converting before multiplication avoids errors in dimensional analysis.\\
      3. \textbf{Knowledge Point}: A narrow strip of constant width covering a certain length can be modeled as a rectangle.\\
      \textbf{Key Considerations}: This applies when the strip has uniform width throughout its entire length; the area represents the total surface covered.
    \end{tcolorbox}

    Please reason step by step, and put your final answer within \textbackslash boxed\{\}.

  \end{augprompt}
  \caption{Example augmented prompt with a partial-solution hint}
  \label{fig:augprompt}
\end{figure}

\subsection{Example Augmented Prompt}
\label{appendix:appendix_augprompt}

This section presents a concrete data example of our augmented prompt format; see Figure~\ref{fig:augprompt}.

\section{Effect of Tolerance Threshold}
\label{appendix:appendix_delta}

\begin{table*}[t]
  \centering
  \scriptsize
  \setlength{\tabcolsep}{3pt}
  \resizebox{\textwidth}{!}{%
    \begin{tabular}{lllllllllll}
      \toprule
      \makecell[l]{$\delta$} & AIME24 & AIME25 & BRUMO25 & \makecell[l]{HMMT\\Feb 25} & AMC23 & CMIMC25 & MATH-500 & \makecell[l]{Olympiad\\Bench} & Avg. & \makecell[l]{Avg.\\\#KP}\\
      \midrule
      0/32 & 63.13 & 49.90 & 64.06 & 34.48 & 91.60 & 33.40 & 92.55 & 73.95 & 62.88 & 1.19 \\
      1/32 & 64.44 & 50.57 & 65.03 & 35.77 & 91.71 & 36.70 & 92.90 & 74.11 & 63.90 & 2.57 \\
      2/32 & 64.05 & 50.30 & 64.80 & 35.20 & 91.33 & 35.90 & 92.85 & 73.70 & 63.52 & 3.45\\
      \bottomrule
  \end{tabular}}
  \caption{Effect of tolerance threshold $\delta$ on offline performance and KP compactness. $\delta=1/32$ provides the best balance between average accuracy and average number of selected KPs.}
  \label{tab:delta_change}
\end{table*}

To set the tolerance threshold in CBRS, we compare $\delta\in\{0/32,1/32,2/32\}$. Figure~\ref{fig:subB} shows that with $\delta=0/32$, the intersection of near-optimal candidates across runs is often too small, making selection brittle; with $\delta=2/32$, the overlap rises to around 60\%, but the selected KP set becomes much larger. As reported in Table~\ref{tab:delta_change}, $\delta=1/32$ achieves the best overall result (highest average accuracy, 63.90) while keeping the KP count moderate (2.57), providing a strong balance between performance and compactness.

\begin{table*}[t]
  \centering
  \setlength{\tabcolsep}{4pt}
  \resizebox{\textwidth}{!}{%
    \begin{tabular}{llllllllll}
      \toprule
      Model & AIME24 & AIME25 & BRUMO25 & \makecell[l]{HMMT25} & AMC23 & CMIMC25 & MATH & \makecell[l]{OlyBench} & Avg. \\
      \midrule
      KnowRL-Nemotron-1.5B & 69.79 & 64.69 & 69.48 & 41.04 & 95.55 & 44.14 & 95.70 & 80.23 & 70.08 \\
      w/o entropy annealing & 68.65 & 62.19 & 67.40 & 39.27 & 95.94 & 42.81 & 94.67 & 77.95 & 68.61 \\
      JustRL & 69.69 & 62.92 & 66.88 & 40.63 & 96.02 & 41.72 & 94.15 & 76.59 & 68.58 \\
      \bottomrule
  \end{tabular}}
  \caption{Ablation of entropy annealing on KnowRL-Nemotron-1.5B across eight evaluation benchmarks.}
  \label{tab:entropy_annealing}
\end{table*}

\end{document}